\documentclass[journal]{IEEEtran}

\usepackage{makecell}
\usepackage{amsmath,amssymb}
\usepackage{graphicx}
\usepackage{setspace}
\usepackage{longtable}
\usepackage{colortbl,booktabs}
\usepackage{tabularx,multirow}
\usepackage{xcolor}
\usepackage{tikz}
\usepackage{multirow}
\usepackage{comment}
\graphicspath{{./pic/}}
\usepackage{color}
\usepackage{algorithm}
\usepackage{algorithmic}
 
\newlength\myindent
\usepackage{mathrsfs} 

\usepackage{array}  
\usepackage{booktabs}

\usepackage{hyperref}
\usetikzlibrary{fadings}
\usetikzlibrary{patterns}
\usetikzlibrary{shadows.blur}
\usetikzlibrary{arrows,shapes,chains}
\usepackage[numbers]{natbib}
\newcommand{\figureref}[1]{Figure~{\ref{#1}}}
\newcommand{\figref}[1]{Fig.~{\ref{#1}}}

\newcommand{\tableref}[1]{Table~{\ref{#1}}}

\makeatletter
\def\hlinewd#1{%
	\noalign{\ifnum0=`}\fi\hrule \@height #1 \futurelet
	\reserved@a\@xhline}
\makeatother

\begin{document}

\title{Semi-Supervised Object Detection with Uncurated Unlabeled Data for Remote Sensing Images}

\IEEEtitleabstractindextext{
\begin{abstract}

Annotating remote sensing images (RSIs) presents a notable challenge due to its labor-intensive nature. Semi-supervised object detection (SSOD) methods tackle this issue by generating pseudo-labels for the unlabeled data, assuming that all classes found in the unlabeled dataset are also represented in the labeled data. However, real-world situations introduce the possibility of out-of-distribution (OOD) samples being mixed with in-distribution (ID) samples within the unlabeled dataset.
In this paper, we delve into techniques for conducting SSOD directly on uncurated unlabeled data, which is termed Open-Set Semi-Supervised Object Detection (OSSOD). Our approach commences by employing labeled in-distribution data to dynamically construct a class-wise feature bank (CFB) that captures features specific to each class. Subsequently, we compare the features of predicted object bounding boxes with the corresponding entries in the CFB to calculate OOD scores. We design an adaptive threshold based on the statistical properties of the CFB, allowing us to filter out OOD samples effectively.
The effectiveness of our proposed method is substantiated through extensive experiments on two widely used remote sensing object detection datasets: DIOR and DOTA. These experiments showcase the superior performance and efficacy of our approach for OSSOD on RSIs.

\end{abstract}

\begin{IEEEkeywords}
Remote sensing images, few-shot learning, meta-learning, object detection, transformation invariance. 
\end{IEEEkeywords}}

\markboth{Journal of \LaTeX\ Class Files,~Vol.~14, No.~8, August~2015}%
{Shell \MakeLowercase{\textit{et al.}}: Bare Demo of IEEEtran.cls for IEEE Journals}
	\author{Nanqing~Liu, Xun~Xu, Yingjie~Gao, Heng-Chao Li
	\IEEEcompsocitemizethanks{
		\IEEEcompsocthanksitem 

        This work was supported in part by the National Natural Science Foundation of China under
        Grant 62271418, and in part by the Natural Science Foundation of Sichuan Province under Grant 23NSFSC0058.

		Nanqing~Liu (lansing163@163.com), and Heng-Chao Li (lihengchao\_78@163.com) are with the School of Information Science and Technology, Southwest Jiaotong University, Chengdu, China. 

		Xun Xu (xux@i2r.astar.edu.sg) is with I2R, A-STAR, Singapore 138632, and also with the School of Electronic and Information Engineering, South China University of Technology, Guangzhou 510640, China.
  
        Yingjie Gao (gaoyingjie@buaa.edu.cn) is with the School of Computer Science and Engineering, Beihang University, Beijing, China.
  
}
	\thanks{Manuscript revised \today.}}
\maketitle
\IEEEdisplaynontitleabstractindextext

\section{Introduction}

Remote sensing images (RSIs) have found a wide range of applications across various fields, including resource surveys, environmental monitoring, and urban planning. In this context, the task of object detection within RSIs plays a crucial role by providing valuable insights for making well-informed decisions. However, current object detection methods based on deep learning, which rely on complete supervision \cite{csff,10005072,9724188,10101785,Liu2021,10198272,glfpn}, often depend on a substantial amount of annotated data. This annotation process can be both time-consuming and financially demanding. Additionally, due to the presence of densely distributed objects and intricate foreground-background variations in RSIs, labeling data introduces significant complexities.

Fortunately, with advancements in remote sensing technology, a vast amount of high-resolution image data has been collected without being annotated. As a result, the utilization of unlabeled data for object detection has become an increasingly important issue that needs to be addressed.

\begin{figure}[t]

	\centering
\resizebox{0.95\linewidth}{!}{\input{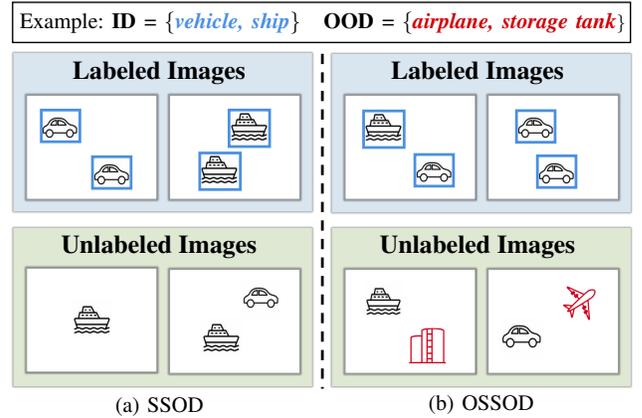}}
	\caption{\footnotesize{
(a) In the context of semi-supervised object detection (SSOD), the underlying assumption is that both labeled and unlabeled data originate from the same label space, constituting in-distribution (ID) samples.
(b) Open-set semi-supervised object detection (OSSOD) accommodates the existence of substantial out-of-distribution (OOD) samples, mirroring the complexities of real-world scenarios.}}
\label{fig:concept}
\end{figure}

Semi-supervised learning has gained significant traction for enhancing object detection by exploiting large unlabeled data. In this context, the utilization of pseudo-label-based semi-supervised object detection (SSOD) methods~\cite{9554202,ssodrsi,hua2023sood} has emerged as a notable approach, yielding promising results in various research studies. This approach operates within a self-training framework, employing two networks simultaneously. It draws upon both labeled data and pseudo-labels generated from unlabeled data to guide the training process. However, these methods inherently assume that the label distributions of both labeled and unlabeled data are identical, as depicted in Fig.~\ref{fig:concept}(a). This assumption implies that the unlabeled data are carefully curated to remove unrelated samples.   

In the context of large-scale remote sensing image (RSI) datasets, practical situations can introduce a notable presence of out-of-distribution (OOD) objects within the unlabeled data. When incorporating OOD data into the self-training phase of SSOD, the conventional practice of treating OOD data as pseudo-labels can inadvertently mislead the network optimization process. Curating such an unlabeled dataset would introduce unwanted additional costs. As a result, we delve into a more intricate and practically significant challenge: Open-Set Semi-Supervised Object Detection (OSSOD) which is able to learn from uncurated unlabeled data, as illustrated in Fig.~\ref{fig:concept}(b).

Addressing this complex challenge involves the incorporation of Out-of-Distribution (OOD) detection mechanisms into the framework of Semi-Supervised Object Detection (SSOD), thereby facilitating the identification of OOD instances during the generation of pseudo-labels.
Recent research~\cite{ossod,wang2023online} has delved into both online and offline OOD detection strategies for OSSOD. Online OOD detection entails the integration of OOD detection algorithms \cite{hendrycksdeep,odin,lee2018simple} as an additional component within the existing object detection algorithm. However, the harmonization of OOD detection and semi-supervised learning presents difficulties that can result in suboptimal performance. In contrast, the offline OOD detection method involves fine-tuning the self-supervised DINO model~\cite{dino} and employing the refined model for region-of-interest (RoI) classification. Nevertheless, the direct application of this offline approach to RSIs yields unsatisfactory results. Several factors contribute to this lackluster performance.
Firstly, the pretraining of the DINO model on natural images limits its adaptability to the nuances of RSIs. Moreover, RSIs exhibit significant variations in scale, while the DINO model necessitates consistent image sizes for accurate classification. These discrepancies undermine the effectiveness of the offline OOD detection method when applied directly to RSIs.

To overcome these limitations, we are inspired by the success of unsupervised anomaly detection~\cite{roth2022towards,li2021cutpaste} and find that utilizing features of labeled in-distribution~(ID) data can serve as a good indicator for OOD detection. Specifically, we first employ a \textbf{class-wise feature bank (CFB)} to store feature representations of each ID category during the training process. To accommodate dynamically updating model weights, we introduce a first-in-first-out~(FIFO) queue to realize the feature bank. Next, we compare the pseudo prediction features with the CFB using K nearest neighbor match for \textbf{OOD score generation}.  Finally, we develop \textbf{adaptive OOD threshold} by considering the distribution of OOD scores within ID samples. This process takes into account the distribution of OOD scores within the ID samples, thereby enhancing the interpretability of our method from a conventional outlier detection perspective. We demonstrate the effectiveness of controlled OSSOD tasks. More importantly, our method exhibits a critical capability in enhancing object detection on fully labeled datasets augmented with uncurated unlabeled data, suggesting its applicability in real-world applications.

Our main contributions can be summarized as:
\begin{itemize}

\item We develop a dynamic class-wise feature bank for pruning OOD unlabeled data, which solely leverages the in-distribution labeled data and does not require additional models for OOD detection, thus having minimal impact on training speed.

\item We design an adaptive method for determining the OOD threshold, which has a nice interpretability from conventional outlier detection perspective.

\item We demonstrate that the proposed method is able to consistently maintain state-of-the-art performance 
 on OSSOD tasks with uncurated unlabeled data on RSIs.
\end{itemize}

\section{Related Works}
\subsection{Label Efficient Object Detection in RSIs}
Object detection in RSIs has achieved remarkable success with fully supervised methods. However, these approaches heavily rely on extensive data annotation. To address this limitation, people have been exploring alternative techniques that require fewer RSI annotations. These techniques include semi-supervised learning \cite{ssodrsi,hua2023sood,10147338}, weakly-supervised learning \cite{rinet,h2rbox,wsodet,h2rboxv2,mol}, few-shot learning \cite{pcnn,tinet,10056362,gfsod,zhang2023few,wang2023few} and active learning \cite{qu2020deep,goupilleau2020active,uehara2019object},. These methods aim to minimize annotation expenses and speed up the labeling process, making them highly valuable for practical applications in remote sensing. For instance, the H2RBox series \cite{h2rbox,h2rboxv2} can detect oriented boxes using only horizontal bounding box annotations, significantly reducing annotation costs. RINet \cite{rinet} and TINet \cite{tinet} focus on extracting more robust features from the perspective of geometric transformation invariance in situations of data scarcity. The datasets used by these methods have been cleaned, and the data either comes from within the same dataset, or from an unlabeled dataset that shares the same categories as the labeled one. In this paper, we explore how to effectively utilize uncurated unlabeled data in RSIs.

\subsection{Semi-Supervised Object Detection}
Semi-supervised learning techniques have emerged as successful methods for image classification by employing data augmentation and consistency regularization on unlabeled data. These techniques have also been extended to object detection tasks, with consistency-based and pseudo-label-based methods being the primary approaches. 
Consistency regularization assumes the manifold or smoothness of the data and encompasses methods where realistic perturbations of the data points do not alter the model output. Conversely, pseudo-labeling methods rely on the high confidence of pseudo-labels and can be added to the training dataset as labeled data. In this paper, we focus on the latter approach. 
Pseudo-label-based methods typically use a student-teacher architecture. For example, STAC \cite{stac} generates pseudo labels by inputting unlabeled images to a teacher model, which are then fed into the model for fine-tuning with strong augmentation. However, this method only generates a pseudo label once, and it will not be updated during training. Instant-Teaching \cite{it} thus solves it by generating pseudo-labels on the fly. 
To simplify the offline pseudo-labeling process and generate more stable pseudo-labels, mean teacher-based methods \cite{ut,st,lm,li2022pseco,wang2023consistent,kar2023revisiting} employ exponential moving average (EMA) to update the teacher network while performing a weak data transformation for online pseudo labeling and a strong data transformation for model training. Several studies \cite{dsl,ut2,dt,et,liu2023ambiguity} then extended these existing methods into one-stage detectors to pursue simple and effective paradigms. More recently, Semi-DETR \cite{semidetr} explored a DETR-based framework for SSOD.

\begin{figure*}[ht]

	\centering
\resizebox{0.95\linewidth}{!}{\input{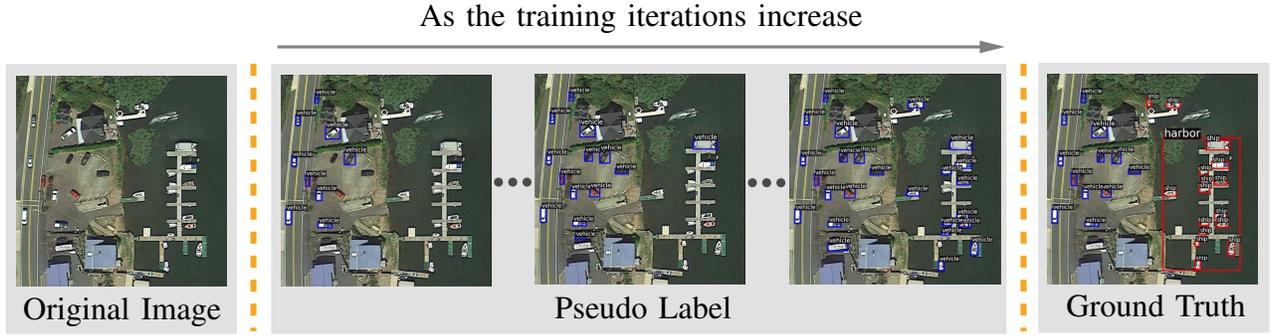}}
	\caption{\footnotesize{Illustration of semantic expansion. In this case, ``vehicle'' is the ID category (depicted in \textcolor{blue}{blue} box), while ``ship'' and ``harbor'' are OOD category (depicted in \textcolor{red}{red} box). Typically, in the closest semi-supervised detector, the training process relies on pseudo-labels. Due to the similarity between some OOD samples and ID samples, they may be mistakenly classified as ID samples. Merely using confidence thresholding based on bounding box scores cannot effectively suppress the appearance of these OOD objects in pseudo-labels. In self-training methods, the use of noisy pseudo-labels exacerbates the open-set problem after several training iterations.}} 

\label{fig:semantic_expansion}
\end{figure*}

\subsection{Open-Set Semi-Supervised Learning}
Open-set semi-supervised learning (OSSL) is a realistic setting of semi-supervised learning where the unlabeled training set contains classes that are not present in the labeled set. Prior works on open-set semi-supervised learning \cite{9956506,mtc,openmatch,chen2020semi,yu2023adaptive,yang2022knowledge,wallin2023improving,li2023iomatch,zhao2023exploration} have primarily focused on image classification tasks. For example, MTC \cite{mtc} utilizes a joint optimization framework to estimate the OOD score of unlabeled images, which is achieved by alternately updating network parameters and estimated scores. OpenMatch\cite{openmatch} applies consistency regularization on a one-vs-all classifier, which serves as an OOD detector to filter the OOD samples during semi-supervised learning. \citet{wallin2023improving} propose an OSSL framework that facilitates learning from all unlabeled data through self-supervision and utilizes an energy-based score to accurately recognize data belonging to the known classes.
Despite these promising results, OSSL for object detection tasks is more challenging than image classification tasks because one image typically contains more instances. In this work, we aim to address open-set semi-supervised object detection by introducing an efficient add-on OOD detection method.

\section{Methodoglogy}
In this section, we first briefly review semi-supervised object detection and then propose an open-set semi-supervised object detection framework.

\subsection{Revisiting Semi-Supervised Object Detection}

Semi-Supervised Object Detection (SSOD) aims to train an object detector using limited labeled data $\mathcal{D}_l=\left\{{x}_i^l, {y}_i^l\right\}_{i=1\cdots N_s}$ and large amount of unlabeled data $\mathcal{D}_u=\left\{{x}_i^u\right\}_{i=1\cdots N_u}$ where the ground-truth label consists of bounding box coordinates and class label ${y}_i^l=\left\{b_{ij},c_{ij}\right\}_{j=1\cdots N_{i}}$. By default, the objects in both labeled and unlabeled sets are drawn from a known category list $\mathcal{Y}=\{1,\cdots, C\}$. 
A typical approach towards semi-supervised object detection~(SSOD) is by self-training on the unlabeled data~\cite{ut,st,lm}. Specifically, self-training experiences two stages of training. In the first stage, a.k.a. the Burn-In stage, the object detection model is trained with all labeled data $\mathcal{D}_l$, resulting in an initial model $\Theta_{init}$. In the second stage, both the teacher model $\Theta_{tea}$ and student model $\Theta_{stu}$ are initialized with $\Theta_{init}$.
The predictions above a confidence threshold $\tau$, made by the teacher model on the unlabeled data, are treated as pseudo labels $\hat{y}_i^u$ (both bounding box and class labels), which are further used for supervising the training of student model. Importantly, the student and teacher models take weakly and strongly augmented images as input respectively to improve generalization. The loss for labeled data $\mathcal{L}_s$ and for unlabeled data $\mathcal{L}_u$ are respectively defined on labeled data and unlabeled data. The final objective is to optimize the weighted sum of both loss terms.

\begin{equation}
\mathcal{L}_{total}=\mathcal{L}_s+\lambda \mathcal{L}_u
\end{equation}

To reduce the negative impact of incorrect pseudo labels, the teacher model is often updated in an exponential moving average manner.
\begin{equation}
\Theta_{tea} \leftarrow \alpha \Theta_{tea}+(1-\alpha) \Theta_{stu},
\end{equation}

\subsection{Open-Set Semi-Supervised Object Detection}

\begin{figure*}[!htp]

	\centering
\resizebox{1.0\linewidth}{!}{\input{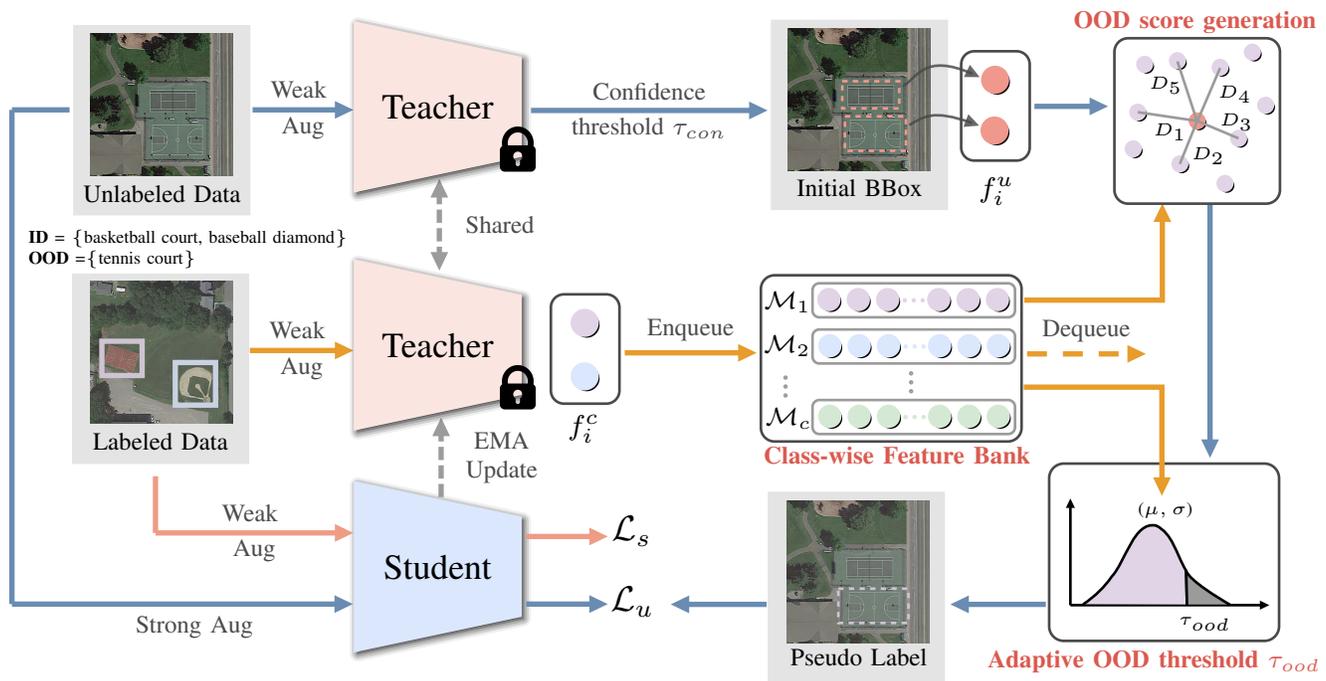}}

	\caption{\footnotesize{Overview of the proposed open-set semi-supervised object detection framework. The entire pipeline's input includes labeled data (containing only ID categories) and unlabeled data (containing either ID or OOD categories). Assuming 'basketball court' and 'baseball diamond' represents the ID category, and 'tennis court' represents the OOD category. In each training iteration, labeled sample feature $f_i^c$ is pushed into the Class-wise Feature Bank (CFB) $\mathcal{M}_c$. Next, the OOD score for $f_i^u$ is computed w.r.t. CFB. Finally, an adaptive OOD threshold $\tau_\text{ood}$ is employed to filter out potential OOD pseudo predictions.}} 

\label{fig:ossod}
\end{figure*}

When unlabeled data are not drawn from the same label space as labeled data, blindly using the out-of-distribution~(OOD) unlabeled data could harm semi-supervised learning. OOD unlabeled objects could be incorrectly classified as in-distribution objects, causing noisy pseudo labels. This phenomenon is also referred to as \textit{semantic expansion}\cite{ossod} (shown in Fig. \ref{fig:semantic_expansion}). This issue becomes more severe when one wishes to push the upper limit of object performance by augmenting existing labeled data with uncurated unlabeled data. 
To minimize the negative impact of OOD unlabeled data, the key challenge lies in successfully pruning out OOD pseudo labels. These OOD detections are substantially harder to differentiate than normal negative anchors because they are often semantically meaningful and visually similar to the ID objects. Setting a hard threshold on classification confidence or regression stability~\cite{st} will very likely fail. Therefore, we introduce a plug-and-play module, consisting of three key components, namely, the \textbf{Class-wise Feature Bank}, the \textbf{OOD Score Generation}, and the \textbf{Adaptive OOD threshold}, to filter out OOD predictions without specifying a fixed threshold. We demonstrate that the proposed method is compatible with major teacher-student semi-supervised object detection methods. The overview of the proposed OSSOD framework is shown in Fig. \ref{fig:ossod}.

\subsubsection{Class-wise Feature Bank}

Detecting OOD predictions can be interpreted as identifying anomalous patterns in an unsupervised fashion. Inspired by the success of anomaly detection by measuring against nominal features~\cite{roth2022towards}, we introduce a memory bank to store the features of in-distribution labeled objects, which are also referred to as prototypes throughout this work. OOD detection is then implemented to measure the similarity of unlabeled sample features to the prototypes. Nevertheless, there are two major concerns in OSSOD that drive us to further improve from the vanilla memory bank adopted in \cite{roth2022towards}. i) As opposed to extracting nominal features with a frozen pre-trained model, the object detection model experiences constant updates through semi-supervised learning on both labeled and unlabeled data. Hence, the memory bank must be updated in a dynamic manner. ii) As the labeled data are highly imbalanced across different classes, a single memory bank will be dominated by the majority class. Taking these two concerns into consideration, we design a Class-wise Feature Bank~(CFB) to dynamically update class-wise prototypes.
Concretely, we adopt a queue-like structure to realize the CFB, denoted as $\mathcal{M}=\{\mathcal{M}_c\}_{c=1\cdots C}$, adhering to a First-In-First-Out (FIFO) principle. The class-wise feature bank is of a fixed length $|\mathcal{M}_c|=L$. The experimental detail for the selection of $L$ can be found in Section \ref{sec:length_exp}. Given RoI features, $f_i^c$, associated with $c$-th class, extracted from ground-truth bounding boxes, we use the following rule to update the CFB, where $\mathcal{M}_c[0]$ refers to the first element in the queue. 

\begin{equation}\label{eq:updating}
\vspace{-0.1cm}
    \mathcal{M}_c=\mathcal{M}_c \bigcup f_i^c,\quad 
    \mathcal{M}_c=\mathcal{M}_c \setminus \mathcal{M}_c[0]
\end{equation}

In a typical two-stage semi-supervised object detection paradigm, the CFB is initialized as an empty set $\mathcal{M}_c=\emptyset$ at the beginning of the second stage. We fill the queue with labeled sample RoI features using the teacher model without dequeue until all $\mathcal{M}_c$ reach the length $L$. Then, we start semi-supervised training with OOD filtering. We summarize the algorithm for class-wise feature bank updating in Alg.~\ref{alg:1}.

\begin{algorithm}
	\renewcommand{\algorithmicrequire}{\textbf{Input:}}
	\renewcommand{\algorithmicensure}{\textbf{Output:}}
	\caption{Class-wise feature bank updating}
	\label{alg:1}
	\begin{algorithmic}[1]
		\REQUIRE Class-wise feature bank $\mathcal{M}=\left\{\mathcal{M}_1, \mathcal{M}_2, \cdots, \mathcal{M}_C\right\}$, GT RoI feature $f_i^c$
		\ENSURE Updated $\mathcal{M}$
		\STATE Initialize all the queue $\mathcal{M}_c$ in $\mathcal{M}$ if it is the first iteration;
            \REPEAT 
            \FOR{each $\mathcal{M}_c \in \mathcal{M}$ } 
            \STATE{Enqueue the GT RoI feature $f_i^c$ to $\mathcal{M}_c$ according to Eq. \ref{eq:updating};} 
            \ENDFOR
            \UNTIL{the length of all queues reaches $L$;}
            \STATE Start training the entire network.

	\end{algorithmic}  
\end{algorithm}

\subsubsection{OOD Score Generation}

We define an OOD score as the measurement for filtering out out-of-distribution pseudo labels. A commonly adopted approach for OOD scoring is by fitting a parametric distribution, e.g. Gaussian distribution, and the likelihood serves as OOD metric~\cite{li2021cutpaste}. Despite enjoying the advantage of being memory efficient, the density-based method has to make an explicit assumption of the distribution. A single Gaussian distribution may not be sufficient to model the large intra-class variations among prototypes. Alternatively, we choose to evaluate the OOD score in a non-parametric manner~\cite{roth2022towards,gu2019statistical}. Specifically, for each pseudo bounding box predicted by the teacher model, we compare the RoI feature $f^u_i$ w.r.t. the prototypes from the $\hat{c}$-th class where $\hat{c}$ is the predicted pseudo-class label. Instead of comparing against the most similar prototype as adopted by~\cite{roth2022towards}, we compare against multiple similar prototypes to avoid the impact of outlier prototypes. Finally, the OOD score is defined as the average cosine distance in Eq.~\ref{eq:oodscore}, where $nnK(f_i^u,\mathcal{M}_c)$ returns the index of the $k$-nearest neighbors of $f_i^u$ among $\mathcal{M}_c$. We use a fixed ratio to estimate $K=r\cdot L$ to avoid hyperparameter tuning.

\begin{equation}\label{eq:oodscore}
    \gamma_{ood}=1-\frac{1}{K}\sum_{k\in nnK(f^u_i;\mathcal{M}_c)} \frac{f_i^{u\top}\mathcal{M}[k]}{||f_i^{u}||\cdot||\mathcal{M}[k]||}
\end{equation}
We summarize the algorithm for OOD score generation in Alg.~\ref{alg:2}. In addition, the experimental detail for the selection of distance metric can be found in Section \ref{sec:knn_distance_exp}.

\begin{algorithm}
	\renewcommand{\algorithmicrequire}{\textbf{Input:}}
	\renewcommand{\algorithmicensure}{\textbf{Output:}}
	\caption{OOD score generation}
	\label{alg:2}
	\begin{algorithmic}[1]
		\REQUIRE Class-wise feature bank $\mathcal{M}=\left\{\mathcal{M}_1, \mathcal{M}_2, \cdots, \mathcal{M}_C\right\}$, predicted RoI feature $f_i^u$
		\ENSURE OOD score $\gamma_{ood}$
            \STATE Fetch the corresponding queue $\mathcal{M}_c$ in $\mathcal{M}$;
            \STATE Find the the index $nnK(f_i^u,\mathcal{M}_c)$ of $k$ nearest distances in $\mathcal{M}_c$ to $f_i^u$;
            \STATE Calculate the OOD score $\gamma_{ood}$ using Eq. \ref{eq:oodscore};
	\end{algorithmic}  
\end{algorithm}

\subsubsection{Adaptive OOD threshold }

OOD detections on the unlabeled data must be excluded from semi-supervised learning to achieve robust performance. Nevertheless, there is no trivial way to determine a fixed threshold for this purpose. An overly aggressive threshold will likely prune out true OOD detections as pseudo labels, but this could compromise semi-supervised learning performance as ID detections could be filtered out. A conservative threshold may harm the performance as true OOD detection could be used for self-training.
To strike a balance we propose an adaptive threshold estimation strategy. 

We propose to estimate an adaptive threshold from the distribution of OOD scores. As we have no access to the labels on unlabeled data indicating true OOD samples, we use the OOD score distribution estimated from in-distribution samples to determine the cut-off threshold.
Similar to the OOD score generation, we also compute the $k$-nearest neighbors for each prototype w.r.t. the rest prototypes. 
\begin{equation}\label{eq:oodscore_proto}
    \gamma_{\text{ood}i}=1-\frac{1}{K}\sum_{k\in nnK(\mathcal{M}_c[i];\{\mathcal{M}_c[j]|j\neq i\})}\frac{\mathcal{M}_c[i]^{\top}\mathcal{M}[k]}{||\mathcal{M}_c[i]||\cdot||\mathcal{M}[k]||}
\end{equation}

This allows us to identify the OOD scores for in-distribution samples. Subsequently, we fit a Gaussian distribution to the OOD scores and compute the mean $\mu_c$ and the standard deviation $\sigma_c$ on $\{\gamma_{\text{ood}i}\}_{i=1\cdots L}$ for each category. The adaptive threshold is chosen as follows:
\begin{equation}
\tau_{\text{ood}_c} = \mu_c + \beta \sigma_c,
\end{equation}
where $\beta$ is a predefined value (e.g., 0, 1, or 2). As the CFB adopts the FIFO update strategy during network training, the values of $\mu_l$ and $\sigma_l$ are also updated at each iteration.
Furthermore, using a fixed $\beta$ value is still suboptimal. A larger $\beta$ value would make the threshold more lenient, potentially introducing unnecessary OOD samples. Conversely, a smaller $\beta$ value would make the threshold more strict, possibly misclassifying ID samples as OOD samples. 
Thus, we extend $\beta$ to a dynamic form that varies with the training epoch $T$:
\begin{equation}
\beta = \beta_{\text{init}} + (\beta_{\text{final}} - \beta_{\text{init}}) * \frac{t}{T}, \quad t \in [0, T],
\end{equation}
where $\beta_{\text{init}}$ and $\beta_{\text{final}}$ are pre-defined initial and final values of $\beta$. It can be observed that in the early stages of network training, when feature learning is not sufficient, using a stricter $\beta$ value benefits generating pseudo-labels. As feature learning becomes more sufficient in the later stages, the threshold is gradually increased.
Finally, we summarize the algorithm for adaptive threshold determination in Alg.~\ref{alg:3}. For experimental details regarding the fixed or adaptive thresholds and how to choose the value of $\beta$, please refer to Section \ref{sec:beta_exp}.

\begin{algorithm}[t]
	\renewcommand{\algorithmicrequire}{\textbf{Input:}}
	\renewcommand{\algorithmicensure}{\textbf{Output:}}
        
	\caption{Adaptive OOD threshold}
	\label{alg:3}
	\begin{algorithmic}[1]
		\REQUIRE Class-wise feature bank $\mathcal{M}=\left\{\mathcal{M}_1, \mathcal{M}_2, \cdots, \mathcal{M}_C\right\}$, CFB length $L$, current epoch $t$, total epoch $T$, $\beta_{\text{init}}$, $\beta_{\text{final}}$ 
		\ENSURE OOD threshold $\tau_\text{ood}$
            \FOR{$\mathcal{M}_c$ in $\mathcal{M}$}
            \STATE Find $k$-nearest neighbors for each prototype $f_i^c\in\mathcal{M}_c$;
            \STATE Compute average OOD score for each prototype $\gamma_{\text{ood}i}$ by Eq.~\ref{eq:oodscore_proto};
            \STATE Update OOD score statistics for in-distribution prototypes $\mu_c=\frac{1}{L}\sum_{i=1}^{L} \gamma_{\text{ood}i}$;
            \STATE $ \sigma_c = \sqrt{\frac{1}{L}\sum_{i=1}^{L}(\gamma_{oodi} - \mu_c)^2}$;
            \STATE $\beta \leftarrow \beta_{\text{init}} + (\beta_{\text{final}} - \beta_{\text{init}}) * \frac{t}{T}$;
            \STATE Update OOD score $\tau_{\text{ood}} = \mu_c + \beta \sigma_c$;

            \ENDFOR
	\end{algorithmic}  
\end{algorithm}

\section{Experiments}

\begin{table*}[t]
    \renewcommand\arraystretch{1.0}
    \caption{Two different class splits on the DIOR dataset.}
    \centering
    \small
    \begin{tabular}{p{2.0cm}<{\centering}|p{7.0cm}<{\centering}|p{7.0cm}<{\centering}}
        \toprule
        &In-distribution classes&Out-of-distribution classes \\
        \midrule
        \multirow{3}*{Split1}
        & chimney, dam, expressway service area, expressway toll station, vehicle, ground track field, overpass, stadium, tennis court, train station& airplane, airport, storage tank, baseball field, golf field, windmill, bridge, ship, basketball court, harbor \\
        \midrule
        \multirow{3}*{Split2} 
        &airplane, airport, storage tank, baseball field, golf field, windmill, bridge, ship, basketball court, harbor & chimney, dam, expressway service area, expressway toll station, vehicle, ground track field, overpass, stadium, tennis court, train station \\

        \bottomrule
    \end{tabular}
    \label{tab:split}
\end{table*}

\subsection{Dataset and Evaluation Protocol}

\begin{figure}[!htb]
    \centering
    \setlength\tabcolsep{2pt} 
    \begin{tabular}{cc}
	\includegraphics[width=0.48\linewidth]{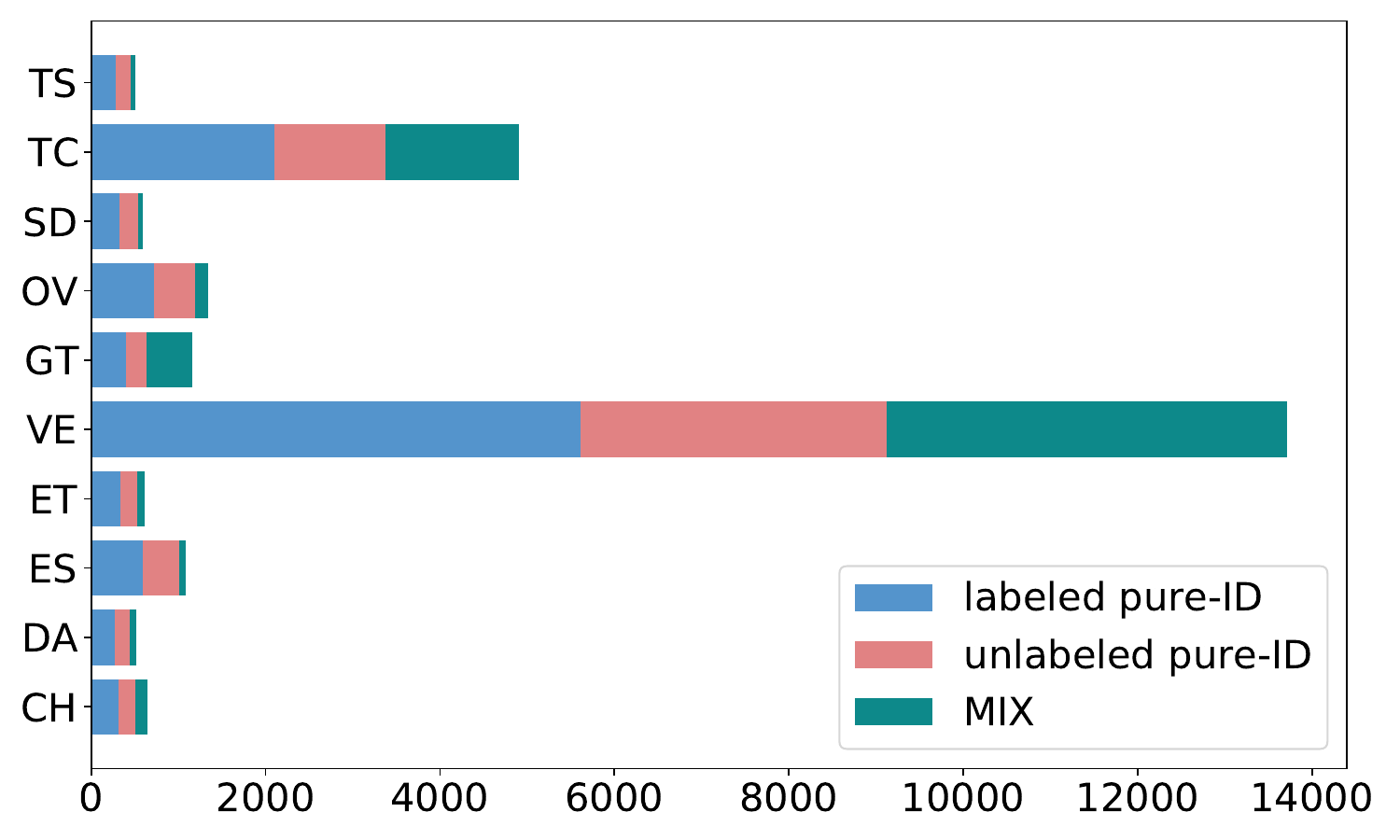}& 
 \includegraphics[width=0.48\linewidth]{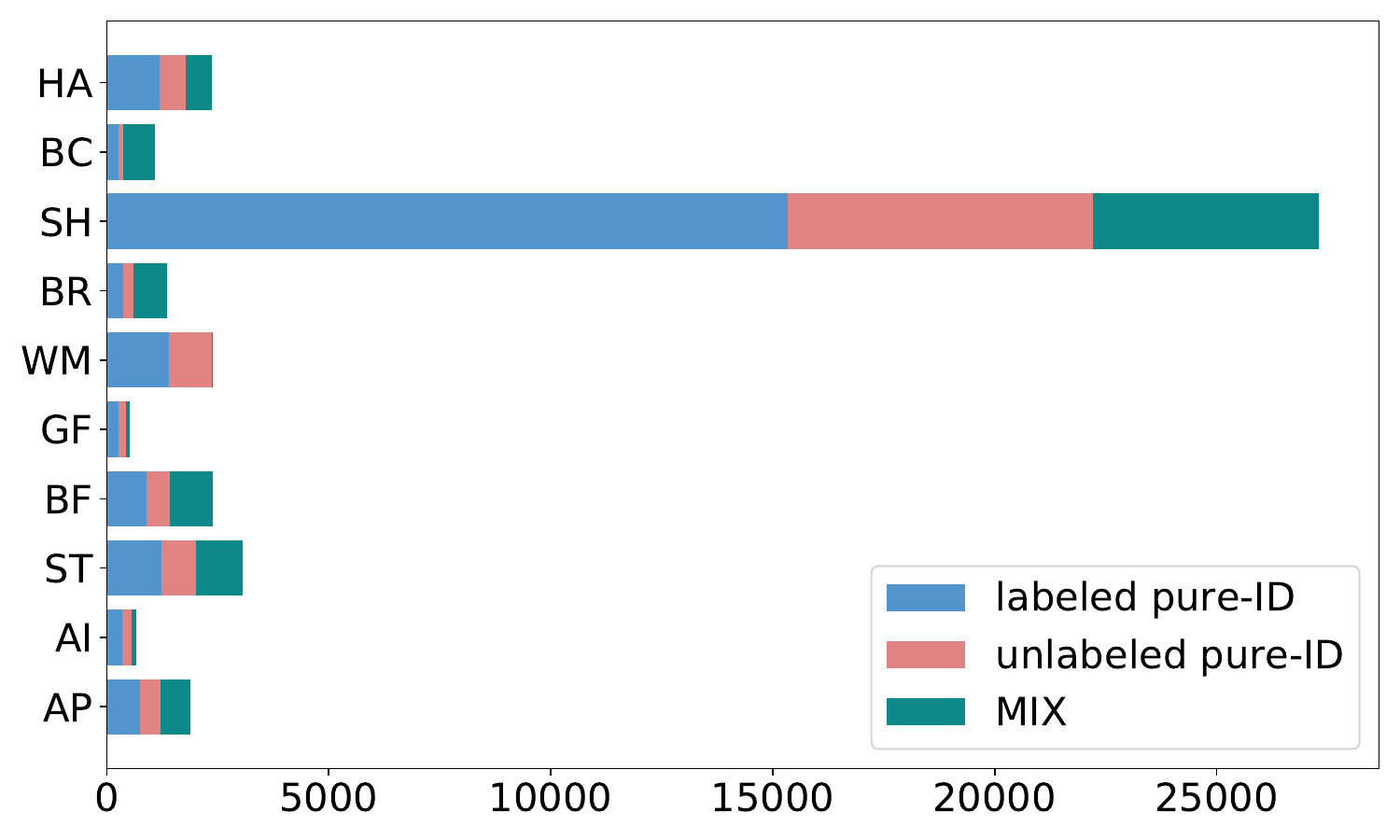}
 \\
   (a) &(b)
    \end{tabular}
    \caption{Number of ID objects in different subsets in Split1 (a) and Split2 (b). The abbreviations for the categories are AP-airplane, AI-airport, BF-baseball field, BC-basketball court, BR-bridge, CH-chimney, DA-dam, ES-expressway service area, ET-expressway toll station, GF-golf field, GT-ground track field, HA-harbour, OV-overpass, SH-ship, SD-stadium, ST-storage tank, TC-tennis court, TS-train station, VE-vehicle, and WM-windmill.}
    \label{fig:object_num}
\end{figure}

To validate the effectiveness of the proposed OSSOD approach, we first carry out a \textbf{Controlled Experiment on DIOR dataset}~\cite{dior}.
The DIOR dataset consists of 23,463 images and 192,472 instances across 20 different object classes. We adopt a similar evaluation protocol as the prior work~\cite{ossod}. Specifically, half classes are selected as in-distribution (ID) classes, and the remaining classes are selected as out-of-distribution (OOD) classes. The specific split of categories is shown in \tableref{tab:split}. The trainval set is split into three subsets: pure-ID, mixed, and pure-OOD image sets. The pure-ID set includes images that have at least one ID object and no OOD objects, while the pure-OOD set comprises images with at least one OOD object and no ID objects. The mixed set includes images with at least one ID object and at least one OOD object. To prevent the inclusion of OOD objects in the labeled set, we randomly sample 3000 pure-ID images as the labeled set and the remaining data as the unlabeled set. 
This process gives rise to four distinct subsets:  labeled \textbf{pure-ID data ($\mathcal{L}$)}, unlabeled \textbf{pure-ID data ($\mathcal{U}$)}, unlabeled \textbf{mixed data ($\mathcal{M}$)}, and unlabeled \textbf{pure-OOD data ($\mathcal{O}$)}.
The number of samples for each ID category in different subsets is shown in the \figureref{fig:object_num}.  It can be observed that each set includes ID samples from every category. 
We also present in \tableref{tab:num_ratio} the details of ID and OOD object distribution under different labeled and unlabeled data schemes, indicating that Split1 is more difficult than Split2.

We further examine the proposed method under a more realistic setting, \textbf{Augmenting Fully Labeled Dataset with Uncurated Unlabeled Data}. Specifically, we augment the fully labeled DIOR dataset with all images from \textbf{DOTA-v1.0}~\cite{dota} as unlabeled data. This dataset consists of images with resolutions ranging from $800 \times 800$ to $4000 \times 4000$. There are approximately 280,000 annotated instances. The training and validation data comprise 1411 images and 458 images, respectively. 
Due to the class mismatch between DIOR and DOTA-v1.0, naively using DOTA-v1.0 as unlabeled data may not fully unleash the power of semi-supervised learning and we demonstrate that the proposed OSSOD can significantly improve the performance with uncurated unlabeled data. We further reverse the role of DIOR and DOTA-v1.0 for more extensive evaluation.  

\begin{table}[!htb]
    \small
    \renewcommand\arraystretch{0.7}
    \caption{The number and ratio of ID and OOD objects vary across different sets in Split1 and Split2}
    \centering
    \setlength\tabcolsep{3pt} 
    \begin{tabular}{l|c|c|c|c}
        \toprule
        &$\mathcal{L}$&$\mathcal{L+U}$&$\mathcal{L+U+M}$&$\mathcal{L+U+M+O}$ \\
        \midrule
        \multicolumn{5}{l}{\emph{Split1}}\\
        \midrule
        Nums of ID&10946&17791&25063&25063\\
        \midrule
        Nums of OOD &0&0&10021&42966\\
        \midrule
        Ratio of OOD&0$\%$&0$\%$&28.6$\%$&63.2$\%$\\
        \midrule
        \multicolumn{5}{l}{\emph{Split2}}\\
        \midrule
        Nums of ID&22076&32945&42966&42966\\
        \midrule
        Nums of OOD&0&0&7272&25063\\
        \midrule
        Ratio of OOD &0$\%$&0$\%$&14.5$\%$&36.8$\%$\\
        \bottomrule
    \end{tabular}
    \label{tab:num_ratio}

\end{table}

\begin{table*}[t]
\renewcommand\arraystretch{1}
\centering
\setlength\tabcolsep{3pt} 
\caption{\footnotesize{Comparison of results from various methods on the DIOR test set in Split1 and Split2.}}
\begin{tabular}{c|l|ccc|ccc|ccc|ccc|c}
\hline
\multirow{2}{*}{\centering Split} & \multirow{2}{*}{\centering Method} & \multicolumn{3}{c|}{$\mathcal{L}$} & \multicolumn{3}{c|}{$\mathcal{L}+\mathcal{U}$} & \multicolumn{3}{c|}{$\mathcal{L}+\mathcal{U}+\mathcal{M}$} & \multicolumn{3}{c|}{$\mathcal{L}+\mathcal{U}+\mathcal{M}+\mathcal{O}$} &\multirow{2}{*}{\begin{tabular}[c]{@{}c@{}}Train Speed  \\(iter/s)\end{tabular}}\\
& & AP & AP$_{50}$ & AP$_{75}$ & AP & AP$_{50}$  & AP$_{75}$ & AP & AP$_{50}$  & AP$_{75}$ & AP & AP$_{50}$  & AP$_{75}$ \\
\midrule
\multirow{7}{*}{1} & \textcolor{gray}{Fully Supervised} & \multirow{7}{*}{35.6} & \multirow{7}{*}{56.6} & \multirow{7}{*}{38.8} & \textcolor{gray}{39.1} & \textcolor{gray}{60.8} & \textcolor{gray}{42.7} & \textcolor{gray}{44.1} & \textcolor{gray}{69.4} & \textcolor{gray}{47.4} & \textcolor{gray}{44.1} & \textcolor{gray}{69.4} & \textcolor{gray}{47.4}&\textcolor{gray}{1.96}\\
& UT \cite{ut} & & & & 38.5 & 60.5 & 41.6 & 37.2 & 57.7 & 40.3 & 34.5 & 53.4 & 37.4 &0.74\\
& UT + MSP \cite{msp}& & & & 38.0 & 60.0 & 40.8 & 37.3 & 58.0 & 40.2 & 35.2 & 54.6 & 38.1 & 0.63 \\
& UT + Entropy & & & & 36.4 & 59.1 & 38.9 & 39.4 & 62.8 & 42.2 & 40.8 & 64.3 & 44.0 & 0.63 \\
& UT + EBM \cite{energy}& & & & 35.7 & 58.2 & 37.9 & 37.0 & 60.1 & 39.3 & 37.4 & 60.9 & 39.8& 0.60\\
& UT + DINO \cite{ossod}& & & & 38.2 & 60.6 & 41.2 & 40.9 & \textbf{64.4} & 40.0 & 41.5 & 64.8 & 45.1&0.32 \\
& UT + Ours & & & & \textbf{39.2} & \textbf{61.1} & \textbf{42.8} & \textbf{41.5} & 64.3 & \textbf{44.6} & \textbf{42.7} & \textbf{65.7} & \textbf{46.1} &0.55\\
\midrule
\multirow{7}{*}{2} & \textcolor{gray}{Fully Supervised} & \multirow{7}{*}{35.1} & \multirow{7}{*}{59.7} & \multirow{7}{*}{36.6} & \textcolor{gray}{38.1} & \textcolor{gray}{62.6} & \textcolor{gray}{40.3} & \textcolor{gray}{42.7} & \textcolor{gray}{68.7} & \textcolor{gray}{45.3} & \textcolor{gray}{42.7} & \textcolor{gray}{68.7} & \textcolor{gray}{45.3}&\textcolor{gray}{1.96} \\
& UT \cite{ut}& & & & 38.5 & 63.9 & 40.3 & 39.1 & 64.0 & 41.4 & 38.2 & 62.6 & 40.2&0.74 \\
& UT + MSP \cite{msp}& & & & 37.2 & 63.6 & 38.3 & 38.3 & 64.7 & 39.3 & 38.5 & 62.7 & 40.9 & 0.63\\
& UT + Entropy & & & & 37.0 & 63.6 & 37.6 & 38.2 & 64.5 & 39.2 & 39.2 & 65.2 & 41.1 & 0.63\\
& UT + EBM \cite{energy}& & & & 35.7 & 62.9 & 35.3 & 36.9 & 64.1 & 37.3 & 37.7 & 64.1 & 39.0 & 0.60\\
& UT + DINO \cite{ossod}& & & & 38.0 & 63.8 & 39.6 & 39.9 & 65.7 & 41.7 & 40.8 & 66.2 & 43.0 &0.32\\
& UT + Ours & & & & \textbf{39.0} & \textbf{64.8} & \textbf{40.5} & \textbf{40.6} & \textbf{66.6} & \textbf{42.6} & \textbf{41.5} & \textbf{67.2} & \textbf{44.0}&0.55 \\
\midrule
\end{tabular}
\label{tab:dior_exp}
\end{table*}

\subsection{Implementation Details}
Our implementation is built upon the MMDetection \cite{mmdetection} codebase. We use Faster-RCNN with FPN and ResNet-50 backbone as the base network. At the beginning of the Burn-In stage, the weights of the feature backbone network are initialized with a pre-trained ImageNet model. We set the OOD score generation ratio $r=1/20$. We use the SGD optimizer with a momentum rate of 0.9 and a learning rate of 0.005, and we employ a constant learning rate scheduler. The batch size is 8. We train for 36 epochs in the Burn-In stage and 12 epochs in the Teacher-Student Mutual Learning stage. We use AP$_{50:95}$ (abbreviated as AP), AP$_{50}$, and AP$_{75}$  as evaluation metrics.

\subsection{Competing Methods}
We evaluate against the state-of-the-art OSSOD method~\cite{ossod} on this task, which consists of both online OOD detection and offline OOD detection methods. For online OOD detection, we compare against \textbf{MSP}~\cite{msp}, \textbf{EBM}~\cite{energy}, and \textbf{Entropy} as competing methods. For MSP, we simply raise the threshold to a higher value. For EBM and Entropy, we add an additional OOD branch on the RoI head. For offline OOD detection, we compare with using \textbf{DINO}~\cite{ossod}. We use ground truth labels to extract patches from cropped images and employ the selective search algorithm to generate background samples with an intersection over union (IoU) less than 0.1 with the ground truth boxes. Finally, we also report the performance of \textbf{Fully Supervised} learning, which treats all data, regardless of whether labeled or not, as labeled data. This serves as the upper bound for competing methods.

\subsubsection{Controlled Experiments on DIOR Dataset}
We first report the results on the two splits on the DIOR dataset in Tab.~\ref{tab:dior_exp}. 
For a fair comparison, we use Faster-RCNN to train on labeled pure-ID data during the Burn-In phase to initialize all competing methods for the second stage semi-supervised training. Unbiased Teacher~(UT)~\cite{ut} is adopted as the base semi-supervised learner.
We make the following observations from the results in Tab.~\ref{tab:dior_exp}. 
\begin{itemize}
    \item All methods start from with in-distribution labeled data ($\mathcal{L}$) only, achieving 35.6\% and 35.1\% AP on Split 1 and 2 respectively. When in-distribution unlabeled data is further included ($\mathcal{L}+\mathcal{U}$), we observe the consistent improvement of \textbf{Fully Supervised} and \textbf{UT}, both of which do not explicitly consider the potential OOD unlabeled data. In comparison, some OSSOD methods, e.g. \textbf{MSP}, \textbf{Entropy} and \textbf{EBM}, underperform UT due to pruning out false negative OOD unlabeled data. Nevertheless, \textbf{Ours} is still able to catch up with \textbf{UT} due to the adaptive thresholding. 
    \item When unlabeled mixed data is further incorporated ($\mathcal{L}+\mathcal{U}+\mathcal{M}$), we observe a significant drop of performance for \textbf{UT} compared with \textbf{Fully Supervised} upper bound~(44.1\% v.s. 37.2\%). We attribute the performance drop to predicting unlabeled OOD samples as in-distribution classes, thus introducing noisy pseudo labels for self-training. In contrast, other OSSOD methods perform generally on par with or better than \textbf{UT}. In particular, \textbf{Ours} achieves substantially better results than \textbf{UT}~(41.5\% v.s. 37.2\%), suggesting filtering out OOD samples is conducive to robust self-training.
    \item With additional unlabeled pure-OOD data ($\mathcal{L}+\mathcal{U}+\mathcal{M}+\mathcal{O}$), we observe a substantial drop of performance for \textbf{UT}~(44.1\% v.s. 34.5\%) due to the negative impact of OOD unlabeled data. Surprisingly, \textbf{UT} even performs worse than \textbf{Fully Supervised} with labeled in-distribution data ($\mathcal{L}$) only, suggesting naive semi-supervised learning could be severely compromised by OOD unlabeled data. In contrast, \textbf{Ours} is able to maintain a competitive performance against all alternative OSSOD methods, suggesting robustness.
    \item Finally, we also compare the training speed. In particular, \textbf{Ours} is only slightly more expensive than \textbf{UT} while achieving far better results when unlabeled data is severely contaminated with OOD samples.
\end{itemize}

\begin{figure}[!htb]

	\centering
	\renewcommand{\tabcolsep}{4pt}
	\begin{tabular}{c}
		\includegraphics[width=0.8\linewidth]{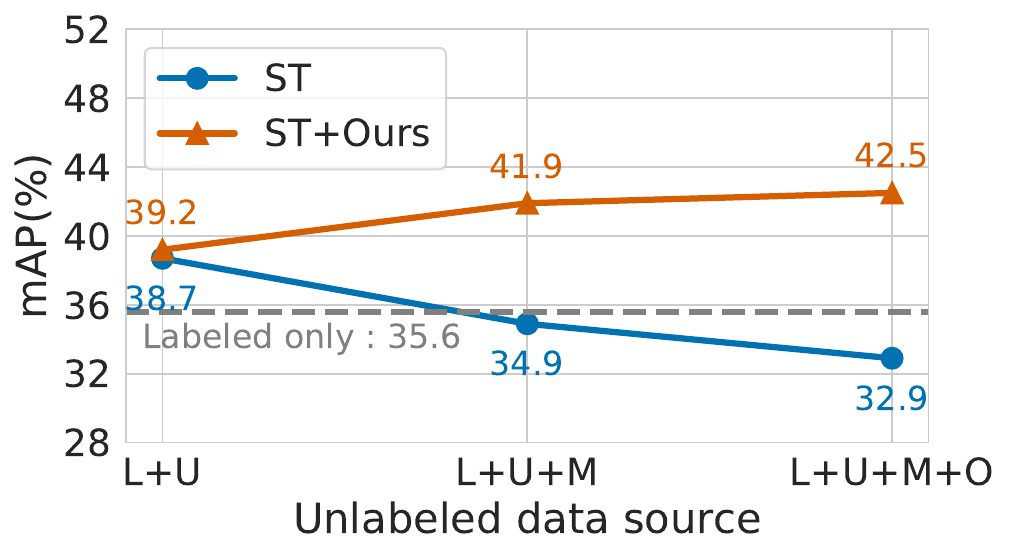} \\
  (a)\\
  \includegraphics[width=0.8\linewidth]{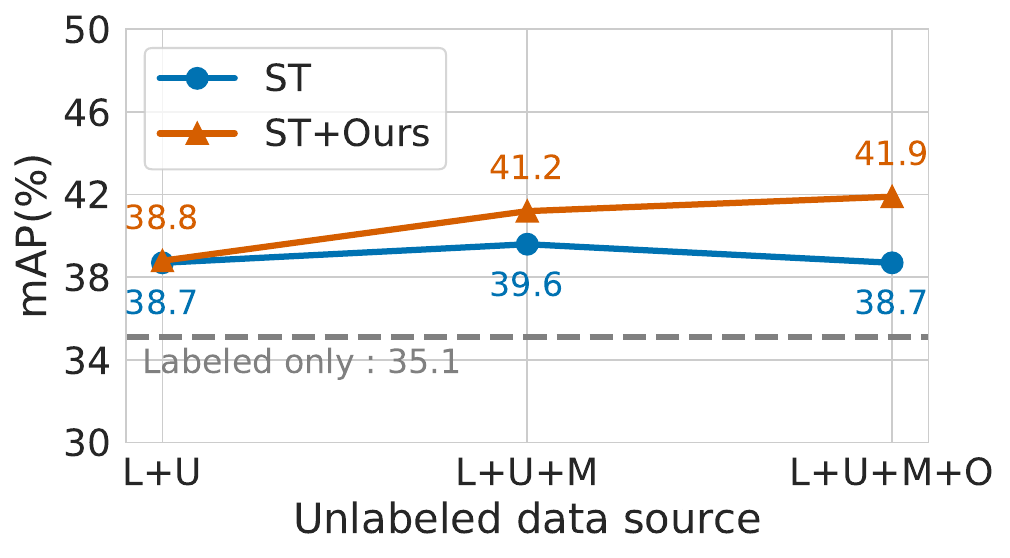}\\
  (b)\\

	\end{tabular}

	\caption{OSSOD results with Soft Teacher (ST) trainer on the DIOR unlabeled dataset Split1 (a) and Split2 (b). Our method consistently improves detection accuracy under OOD-contaminated unlabeled data.}
	\label{fig:st_exp}
\end{figure}

We further validate the effectiveness of our method on an alternative SSOD strategy, Soft Teacher~(ST) \cite{st}. As shown in Fig.~\ref{fig:st_exp}, similar patterns are observed with ST as a semi-supervised trainer. When unlabeled mixed data and unlabeled pure-OOD data are incrementally incorporated, the performance of ST decreases. In contrast, combining ST with our OOD detector can effectively alleviate the negative impact of OOD data. 

\subsubsection{Analysis for the State-of-the-Art Method}
\begin{figure}[!htb]
	
	\centering

	\includegraphics[width=0.9\linewidth]{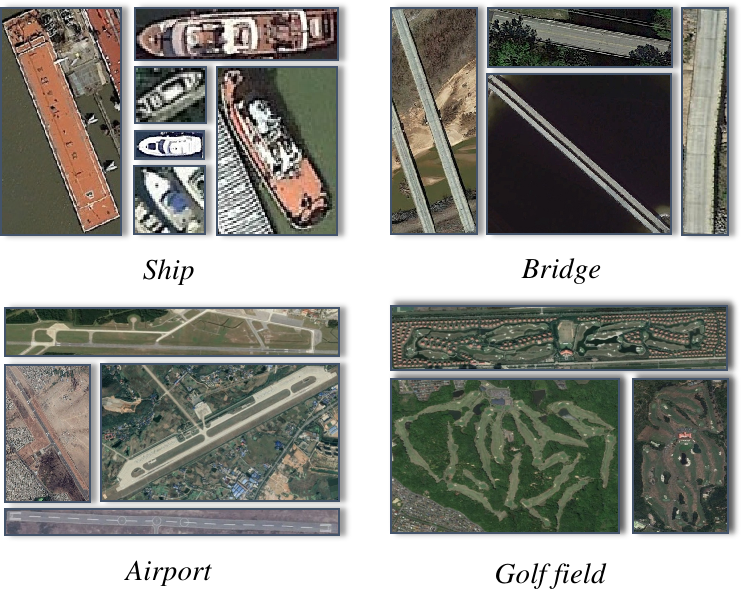}	

	\caption{Cropped images according to the GT boxes.}
	\label{fig:crop}

\end{figure}

\begin{figure}[!htb]
	
	\centering
 	\begin{tabular}{cc}
  
	\includegraphics[width=0.4\linewidth]{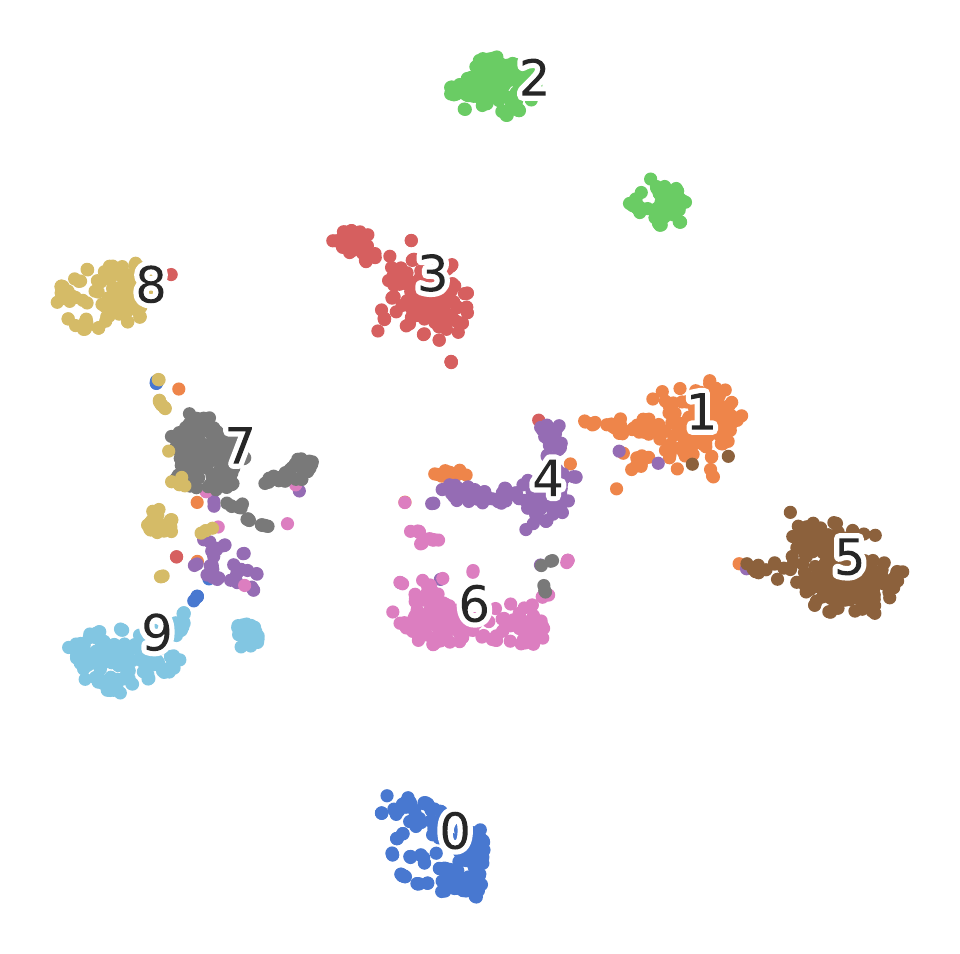}&	\includegraphics[width=0.4\linewidth]{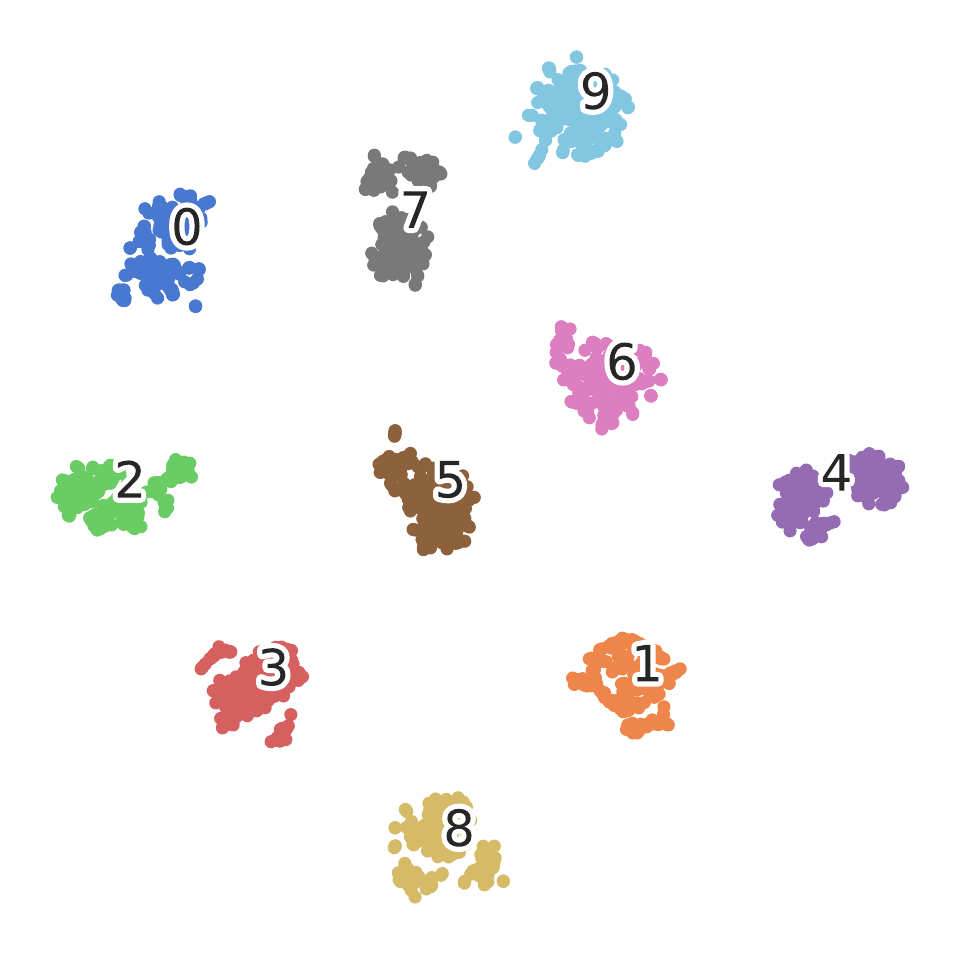}\\
        (a) DINO &(b) CFB\\
	\end{tabular}
	\caption{The t-SNE visualization of the features of DINO(a) and CFB(b) on the ID categories using the DIOR dataset. The abbreviations for the categories are 0-airplane, 1-airport, 2-storage tank, 3-baseball field,
    4-golf field, 5-windmill, 6-bridge, 7-ship, 8-basketball court, 9-harbor.}
	\label{fig:t-SNE}
\end{figure}

We also analyzed why the state-of-the-art OSSOD method (DINO)\cite{ossod} is not suitable for remote sensing images (RSIs). Firstly, the sizes of objects in RSIs vary significantly after cropping, whereas the DINO method has a fixed input size as a classifier. We showcase some patches cropped according to the ground truth bounding boxes from selected categories in \figref{fig:crop}. Secondly, the DINO method is pre-trained on natural images and does not adapt well during fine-tuning on RSIs. As illustrated in the \figref{fig:t-SNE}, the features extracted by DINO exhibit confusion among certain categories. In contrast, directly extracting corresponding features from the network as CFB exhibits strong discriminative ability.

\begin{table}[ht]
    \small
    \caption{\footnotesize{Results on augmenting fully labeled dataset with uncurated unlabeled dataset.}}
    \centering
    \setlength{\tabcolsep}{3pt} 
    \begin{tabular}{l|cc|ccc} 
        \toprule
        Method & Labeled & Unlabeled & AP & AP$_{50}$  & AP$_{75}$  \\
        \midrule
        Fully Supervised & DIOR & - & 45.7 & 71.3 & 49.4 \\
        \midrule
        UT \cite{ut}& DIOR & DOTA & 48.3 & 74.2 & 52.6 \\
        \rowcolor{gray!15}
        UT+Ours & DIOR & DOTA & 49.5 & 75.6 & 54.3 \\
        \midrule
        ST \cite{st}& DIOR & DOTA & 48.8 & 75.4 & 53.4 \\
        \rowcolor{gray!15}
        ST+Ours & DIOR & DOTA & 50.5 & 76.7 & 55.6 \\
        \midrule
        \midrule
        Fully Supervised & DOTA & - & 40.1 & 65.3 & 42.8 \\
        \midrule
        UT \cite{ut}& DOTA & DIOR & 42.0 & 68.0 & 45.0 \\
        \rowcolor{gray!15}
        UT+Ours & DOTA & DIOR & 42.9 & 68.8 & 46.1 \\
        \midrule
        ST \cite{st}& DOTA & DIOR & 41.9 & 68.3 & 44.1 \\
        \rowcolor{gray!15}
        ST+Ours & DOTA & DIOR & 43.2 & 69.8 &46.2 \\
        \bottomrule
    \end{tabular}
    \label{tab:dota_exp}

\end{table}

\begin{figure*}[!htb]

	\centering
	\includegraphics[width=1.0\linewidth]{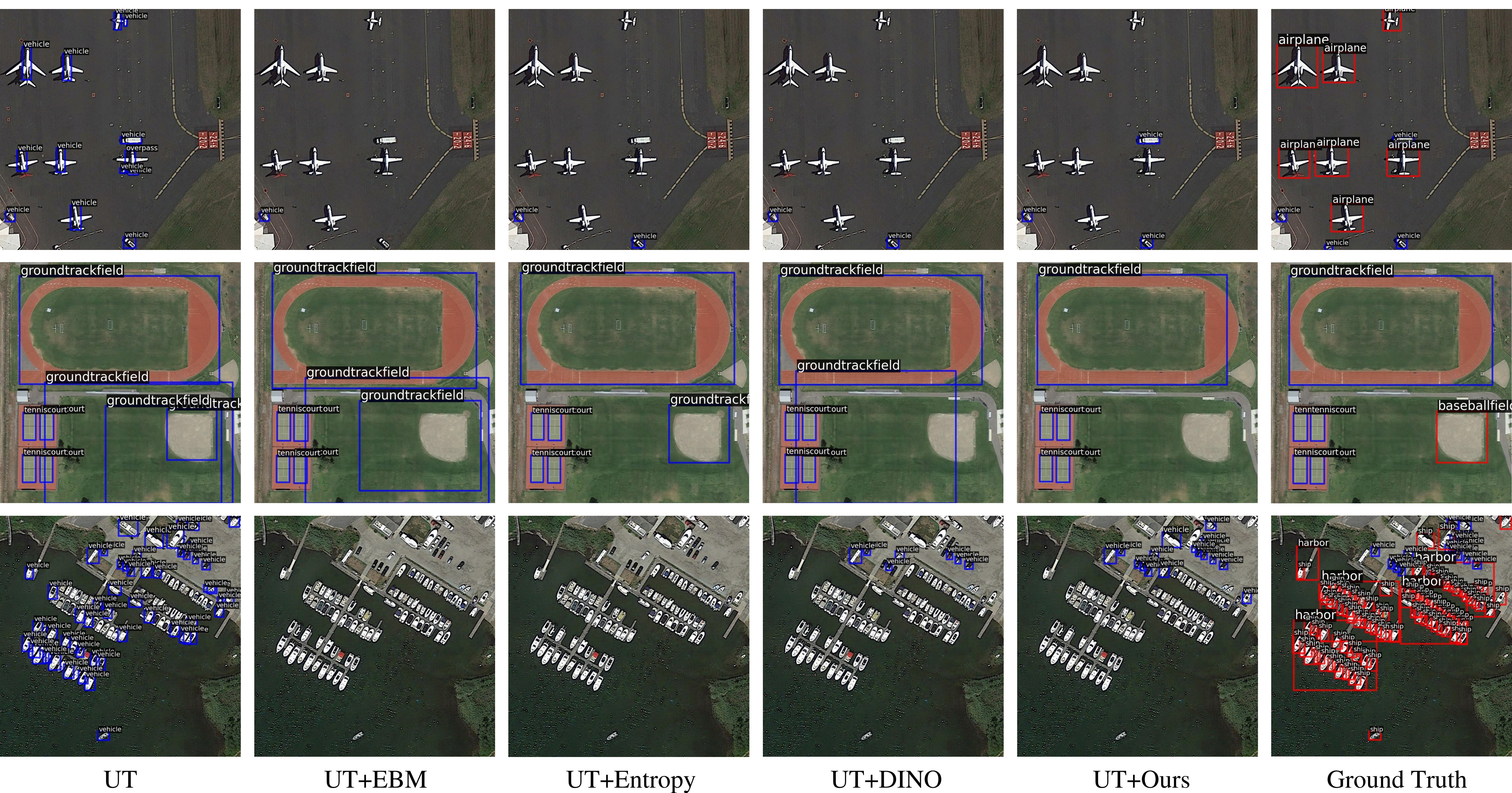}
	

	\caption{Qualitative evaluation of pseudo predictions generated by different methods on DIOR unlabeled training dataset on \textbf{Split1}. Note that the airplane, baseball field, harbor, and ship are OOD categories (indicate with \textcolor{red}{red} boxes in the ``Ground Truth'' view). In-distribution pseudo bounding boxes are colored with \textcolor{blue}{blue} contour.}
    \label{fig:results_comparison_split1}

\end{figure*}

\begin{figure*}[!htb]

	\centering
	\includegraphics[width=1.0\linewidth]{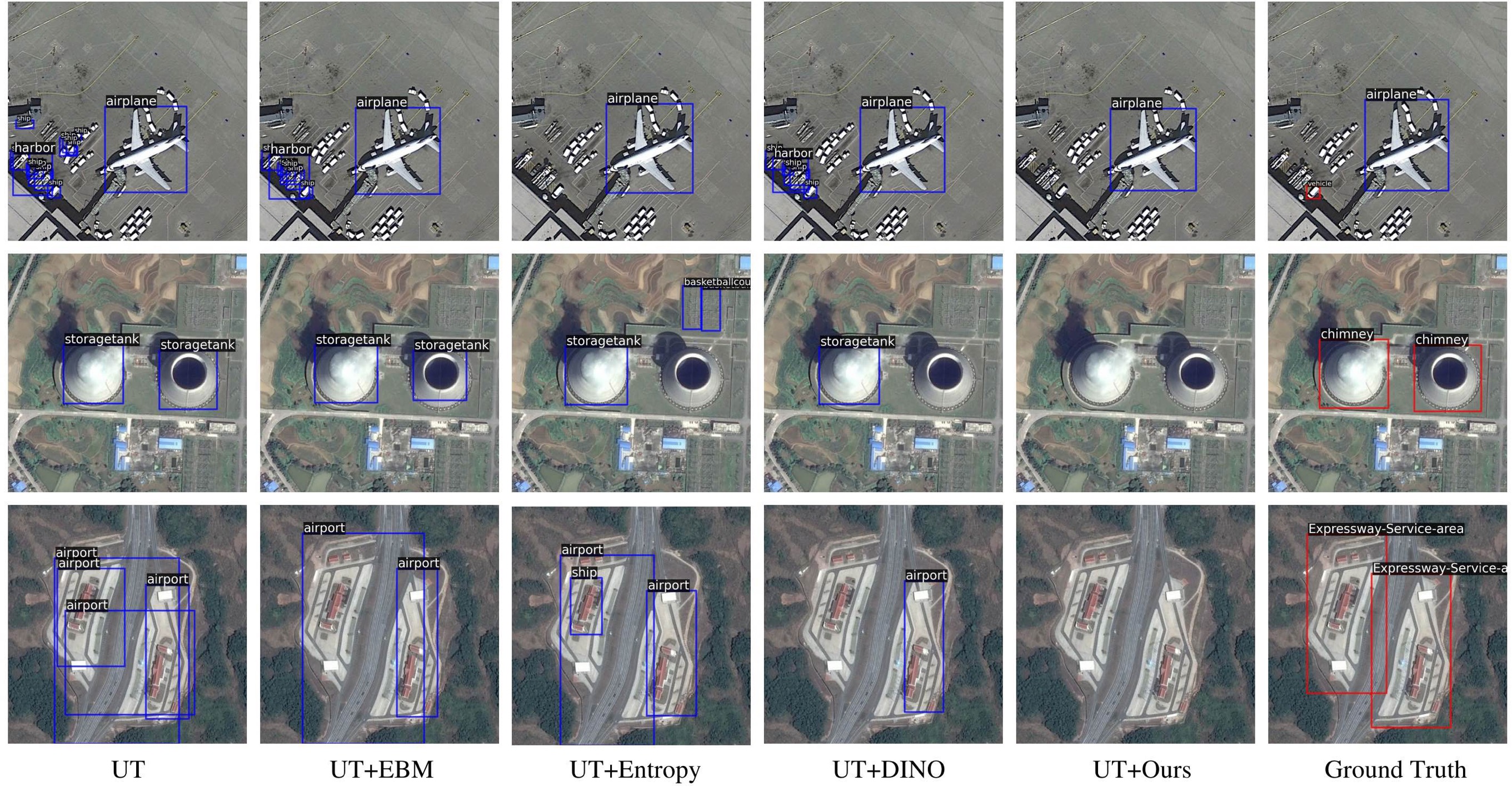}
	

	\caption{Qualitative evaluation of pseudo predictions generated by different methods on DIOR unlabeled training dataset on \textbf{Split2}. Note that the vehicle, Expressway-Service-area, and chimney are OOD categories (indicate with \textcolor{red}{red} boxes in the ``Ground Truth'' view). In-distribution pseudo bounding boxes are colored with \textcolor{blue}{blue} contour.}
    \label{fig:results_comparison_split2}

\end{figure*}

\subsubsection{Augmenting Fully Labeled Dataset with Uncurated Unlabeled Data}
In this section, we present a more realistic open-set semi-supervised object detection experiment setting by augmenting fully labeled data with the uncurated unlabeled dataset. 
 As shown in \tableref{tab:dota_exp}, we first observe that SSOD methods improve over the Fully Supervised baseline with additional uncurated unlabeled data. Nevertheless, the class mismatch between the DIOR and DOTA gives us the opportunity to further boost the performance. With additional OOD detection, the performance on labeled datasets is consistently improved for both \textbf{UT} and \textbf{ST} semi-supervised trainer. This experiment provides a new opportunity for adopting semi-supervised learning with totally uncurated unlabeled data.

\subsubsection{Qualitative Evaluation of Pseudo Predictions}
To more intuitively show the effectiveness of our method, we visualize in Fig.~\ref{fig:results_comparison_split1} and Fig.~\ref{fig:results_comparison_split2}, the generated pseudo bounding boxes of different methods on the DIOR unlabeled training dataset on Split1 and Split2. In Fig.~\ref{fig:results_comparison_split1}, on the first row, we specify ``\textit{airplane}'' (red boxes in ground-truth ) as OOD samples, it can be observed that vanilla \textbf{UT} incorrectly detect airplanes as ID objects (``\textit{overpass}'' and ``\textit{vehicle}''). Other competing methods also exhibit a higher false detection rate compared to our approach.
For more challenging examples, such as ``\textit{vehicle}'' and ``\textit{ships}'' (third row), our method may still exhibit inevitable errors, nevertheless, we still outperform other competing methods. In Fig.~\ref{fig:results_comparison_split2}, the ``\textit{vehicle}'', ``\textit{Expressway-Service-area}'', and ``\textit{chimney}'' are OOD categories. In the first row of images, it can be observed that while all methods are able to effectively recognize ``\textit{airplanes}'', most of them misclassify  ``\textit{vehicle}'' and the surrounding background as ``\textit{harbor}'' except for ours and entropy. For ``\textit{Expressway-Service-area}'' and ``\textit{chimney}'', our method is able to effectively classify them as background. However, other methods struggle to completely differentiate them from ``\textit{airport}'' and ``\textit{storage tank}'', respectively. 

\subsection{Ablation study}

We conduct ablation experiments on DIOR in Split1 to investigate the effectiveness of individual components of the proposed method. We use the Unbiased Teacher \cite{ut} as the semi-supervised trainer.
\subsubsection{Updating Strategy in Class-wise Feature Bank}
We first investigate alternative designs of CFB. Specifically, we compared two update strategies, namely ``Static'' and ``Dynamic'', where ``Static'' freezes the CFB after the Burn-In stage while ``Dynamic'' maintains the CFB as a FIFO queue. As shown in \tableref{tab:dynamic_exp}, ``Dynamic'' update is consistently better than freezing CFB after the Burn-In stage because as training goes the feature representation also experiences constant shift, and updating CFB can always track the feature shift.


\begin{table}[!htb]

	\caption{\footnotesize{Comparing different updating methods of CFB on DIOR test set in Split1.}}
	\centering
 \setlength\tabcolsep{3pt} 
        {
	\begin{tabular}{c|ccc|ccc} 
		\toprule
            \multirow{2}{*}{Updating Method}&\multicolumn{3}{c|}{$\mathcal{L}+\mathcal{U}+\mathcal{M}$}&\multicolumn{3}{c}{$\mathcal{L}+\mathcal{U}+\mathcal{M}+\mathcal{O}$}\\
             &AP & AP$_{50}$  & AP$_{75}$ &AP & AP$_{50}$  & AP$_{75}$\\
            \midrule

		Static&38.3&61.4&40.8&37.7&60.8&40.0\\
            Dynamic&\textbf{41.5}&\textbf{64.3}&\textbf{44.6}&\textbf{42.7}&\textbf{65.7}&\textbf{46.1}\\
            \bottomrule
	\end{tabular}}
    \label{tab:dynamic_exp}
\end{table}

\subsubsection{Class-wise Feature Bank Length $L$}
\label{sec:length_exp}
As illustrated in Table \ref{tab:l_exp}, when the length is set to 100, all performance metrics obtained the best results except AP$_{50}$  ($\mathcal{L}+\mathcal{U}+\mathcal{M}+\mathcal{O}$). The longer the length of CFB, the more features it encompasses. However, at the same time, the probability of incorporating outdated features also increases.

\begin{table}[!htb]

	\caption{\footnotesize{Comparing different $L$ of CFB on DIOR test set in Split1.}}
	\centering
 \setlength\tabcolsep{4pt} 
        {
	\begin{tabular}{c|ccc|ccc} 
		\toprule
            \multirow{2}{*}{$L$}&\multicolumn{3}{c|}{$\mathcal{L}+\mathcal{U}+\mathcal{M}$}&\multicolumn{3}{c}{$\mathcal{L}+\mathcal{U}+\mathcal{M}+\mathcal{O}$}\\
             &AP & AP$_{50}$  & AP$_{75}$ &AP & AP$_{50}$  & AP$_{75}$\\
		\midrule

		20&41.0&63.9&44.3&41.6&64.7&45.3\\
  		50&41.2&64.2&44.4&41.9&65.0&45.5\\
            100&\textbf{41.5}&\textbf{64.3}&\textbf{44.6}&\textbf{42.7}&{65.7}&\textbf{46.1}\\
  		200&41.3&63.6&44.0&42.3&65.4&45.8\\  
            500&40.9&63.5&44.0&42.5&\textbf{65.8}&45.6\\  
		\bottomrule
	\end{tabular}}
    \label{tab:l_exp}

\end{table}

\subsubsection{Distance Metric in $k$-NN}
\label{sec:knn_distance_exp}
As shown in \tableref{tab:distance_exp}, cosine similarity is more suitable as a distance metric in our case.
The input features are obtained from the FC layer in the box head. Based on our observations, these features are high-dimensional and sparse. In this scenario, there is only a marginal difference between using $L_1$ and $L_2$ distances. The cosine similarity ($Cos$), on the other hand, is more effective in distinguishing differences in direction between features while being insensitive to their absolute values. 

\begin{table}[!htb]

	\caption{Comparing different distance metrics of $k$-NN on DIOR test set in Split1.}
	\centering
 \setlength\tabcolsep{5pt} 
        {
	\begin{tabular}{c|ccc|ccc} 
		\toprule
            \multirow{2}{*}{Distance}&\multicolumn{3}{c|}{$\mathcal{L}+\mathcal{U}+\mathcal{M}$}&\multicolumn{3}{c}{$\mathcal{L}+\mathcal{U}+\mathcal{M}+\mathcal{O}$}\\
             &AP & AP$_{50}$  & AP$_{75}$ &AP & AP$_{50}$  & AP$_{75}$\\
		\midrule
  		$L_1$&40.8&62.6&44.3&41.9&64.1&45.3\\
  		$L_2$&40.8&63.0&44.0&42.1&64.8&45.7\\   
		Cosine&\textbf{41.5}&\textbf{64.3}&\textbf{44.6}&\textbf{42.7}&\textbf{65.7}&\textbf{46.1}\\
		\bottomrule
	\end{tabular}}
    \label{tab:distance_exp}
\end{table}

\subsubsection{Adaptive OOD Threshold Updating}
\label{sec:beta_exp}

\begin{figure}[!htb]
	
	\centering

	\includegraphics[width=0.9\linewidth]{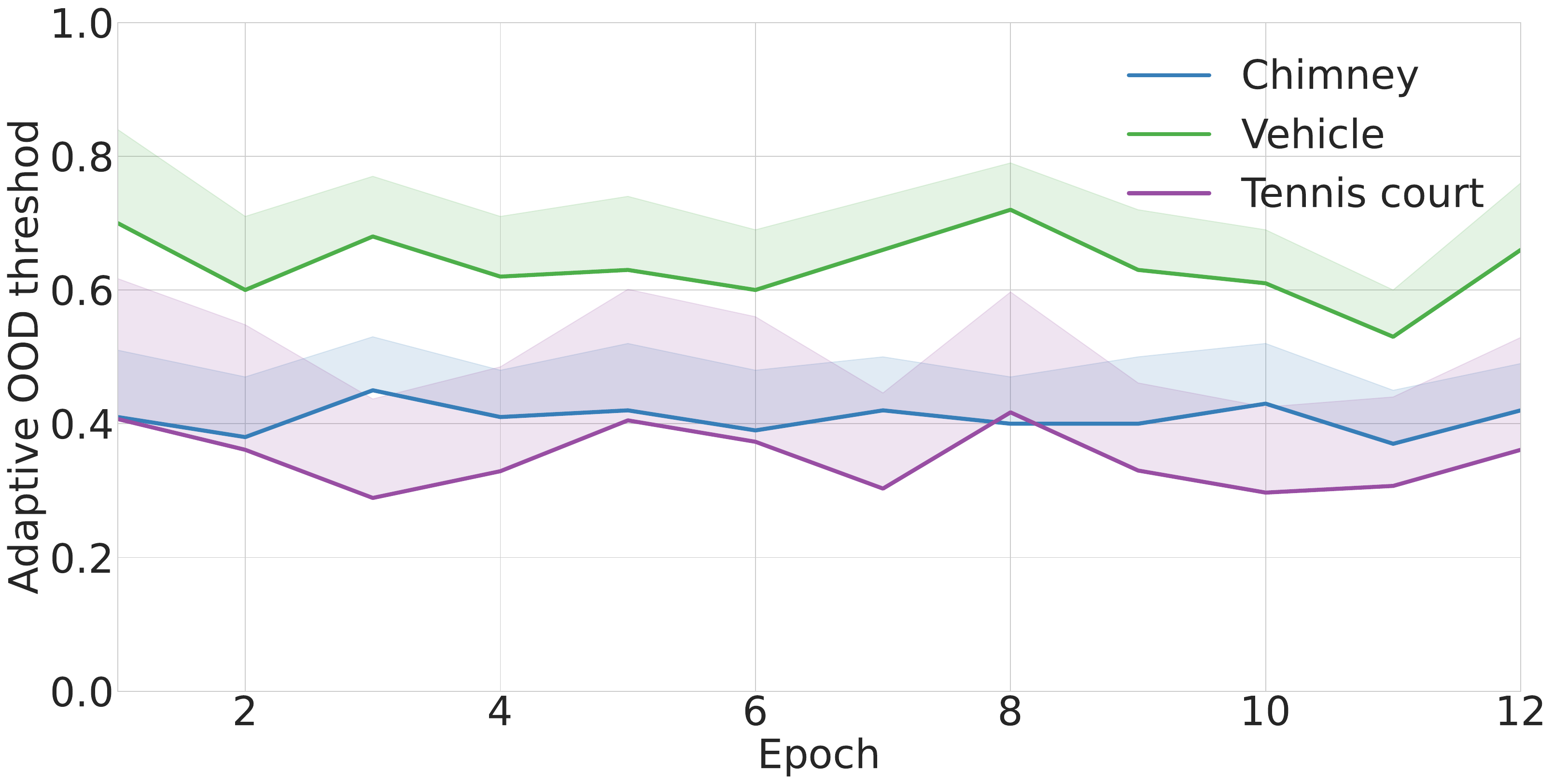}	

	\caption{Adaptive OOD threshold}
	\label{fig:threshold_line}
 \end{figure}

As shown in \tableref{tab:beta_exp}, we compare our proposed adaptive threshold with the fixed threshold approach. For the fixed threshold, we experiment with fixed values of 0.5, 0.6, and 0.7 for each category. It's noticeable that manually chosen thresholds led to suboptimal results and exhibited significant variations as the thresholds changed. In contrast, our method with fixed $\beta$ demonstrated better results than the fixed threshold approach. Additionally, when subjecting $\beta$ to a linear transformation across training epochs, $\beta\in[1,2]$, the results are further improved. We demonstrated the variations of thresholds for different categories as training epochs progressed (shown in \figref{fig:threshold_line}). It can be observed that the values (mean) and ranges of thresholds (standard deviation) for different categories are distinct and a fixed threshold to reliably filter out all OOD predictions does not exist.

\begin{table}[!htb]
    \small
	\caption{Comparing fixed and adaptive OOD threshold on DIOR test set in Split1.}

	\centering
 \setlength\tabcolsep{3pt} 

        {
	\begin{tabular}{c|ccc|ccc} 
		\toprule
            \multirow{2}{*}{$\tau_{\text{ood}}$}&\multicolumn{3}{c|}{$\mathcal{L}+\mathcal{U}+\mathcal{M}$}&\multicolumn{3}{c}{$\mathcal{L}+\mathcal{U}+\mathcal{M}+\mathcal{O}$}\\
             &AP & AP$_{50}$  & AP$_{75}$ &AP & AP$_{50}$  & AP$_{75}$\\
		\midrule
            \multicolumn{6}{l}{\emph{Fixed OOD threshod}}\\
            0.4 & 39.2 & 62.1 & 41.9 & 41.0 & 64.1 & 44.1 \\
		0.5 &40.0&62.5&42.6&41.4&64.8&44.5\\
  		0.6 &40.3&62.9&43.5&40.8&63.5&43.8\\
  		0.7 &38.7&60.4&41.9&38.5&59.4&41.8\\            
            \midrule
            \multicolumn{6}{l}{\emph{Adaptive OOD threshod}}\\
		$\beta$ = 0&40.5&63.3&43.5&41.5&64.4&44.8\\
  		$\beta$ = 1&40.9&63.1&44.2&41.9&64.8&44.8\\
  		$\beta$ = 2&39.7&61.1&43.0&40.7&62.1&44.0\\
      	$\beta \in [1, 2]$&\textbf{41.5}&\textbf{64.3}&\textbf{44.6}&\textbf{42.7}&\textbf{65.7}&\textbf{46.1}\\
      	$\beta \in [0, 2]$&41.0&63.6&44.2&42.2&65.0&45.6\\  
		\bottomrule
	\end{tabular}}
    \label{tab:beta_exp}
\end{table}




\section{Conclusion}
In this paper, we propose an end-to-end open-set semi-supervised object detection method to effectively utilize extensive uncurated data in remote sensing images. We introduce a dynamically updated class-wise feature bank to model the in-distribution samples. A thresholding strategy inspired by a statistical point of view is introduced to adaptively determine OOD predictions. We create OSSOD benchmarks on two widely adopted RSI datasets to showcase the effectiveness of the proposed method. Importantly, we demonstrate that the proposed method enables semi-supervised object detection with uncurated unlabeled data. 

\small
\bibliography{ref/ref}	
\end{document}